\documentclass[remotesensing,article,accept,moreauthors,pdftex]{definitions/mdpi}

\usepackage[flushleft]{threeparttable}
\usepackage{listings}
\usepackage{amssymb}
\usepackage{gensymb}
\usepackage{xcolor}
\newcommand{\R}{\textbf{R}\:}
\newcommand{\sits}{\texttt{sits}\:}
\usepackage{microtype}
\usepackage[all,defaultlines=3]{nowidow}

\newcommand{\codeavailability}[1]{
\vspace{6pt}\noindent{\fontsize{9}{11.2}\selectfont\textbf{Code Availability Statement:} {#1}\par}}

\firstpage{1} 
\pubvolume{13}
\issuenum{1}
\articlenumber{2428}
\pubyear{2021}
\copyrightyear{2021}
\externaleditor{Academic Editor: Pieter Kempeneers} 
\datereceived{29 April 2021} 
\dateaccepted{09 June 2021} 
\datepublished{22 June 2021} 
\hreflink{https://doi.org/10.3390/rs13132428} 

\Title{Satellite Image Time Series Analysis for Big Earth Observation~Data}
\TitleCitation{Satellite Image Time Series Analysis for Big Earth Observation Data}

\Author{Rolf Simoes $^{1,}$*\orcidA{}, 
        Gilberto Camara $^{1}$\orcidB{},
        Gilberto Queiroz $^{1}$\orcidC{},
        Felipe Souza $^{1}$, 
        Pedro R. Andrade $^{1}$\orcidD{}, \linebreak
        Lorena Santos $^{1}$\orcidE{},
        Alexandre Carvalho $^{2}$\orcidF{} and 
        Karine Ferreira $^{1}$\orcidG{}
        }

\AuthorNames{Rolf Simoes, Gilberto Camara, Gilberto Queiroz, Felipe Souza, Pedro R. Andrade,  Lorena Santos, Alexandre Carvalho and Karine Ferreira}

\AuthorCitation{Simoes, R.; Camara, G.; Queiroz, G.; Souza, F.; Andrade, P.R.; Santos, L.; Carvalho, A.; Ferreira, K.}

\address{
$^{1}$ \quad National Institute for Space Research  (INPE), Avenida dos Astronautas, 1758, Jardim da Granja, \linebreak Sao Jose dos Campos, SP 12227-010, Brazil; gilberto.camara@inpe.br (G.C.); gilberto.queiroz@inpe.br (G.Q.); felipe.souza@inpe.br (F.S.); pedro.andrade@inpe.br (P.R.A.); lorena.santos@inpe.br (L.S.);  karine.ferreira@inpe.br (K.F.)\\
$^{2}$ \quad National Institute for Applied Economics Research, SBS, Quadra 1 Bloco J, Brasília, DF 70076-900, Brazil; alexandre.ywata@ipea.gov.br
}

\corres{Correspondence: rolf.simoes@inpe.br}

\abstract{The development of analytical software for big Earth observation data faces several challenges. Designers need to balance between conflicting factors. Solutions that are efficient for specific hardware architectures can not be used in other environments. Packages that work on generic hardware and open standards will not have the same performance as dedicated solutions. Software that assumes that its users are computer programmers are flexible but may be difficult to learn for a wide audience. This paper describes \texttt{sits}, an open-source R package for satellite image time series analysis using machine learning. To allow experts to use satellite imagery to the fullest extent, \sits adopts a time-first, space-later approach. It supports the complete cycle of data analysis for land classification. Its API provides a simple but powerful set of functions. The software works in different cloud computing environments. Satellite image time series are input to machine learning classifiers, and the results are post-processed using spatial smoothing. Since machine learning methods need accurate training data, \sits includes methods for quality assessment of training samples. The software also provides methods for validation and accuracy measurement. The package thus comprises a production environment for big EO data analysis. We show that this approach produces high accuracy for land use and land cover maps through a case study in the Cerrado biome, one of the world’s fast moving agricultural frontiers for the year 2018.}

\begin{document}

\section{Introduction}

The growing demand for natural resources has caused major environmental impacts and is changing landscapes everywhere. Conversion of land cover due to human use is one of the key factors behind greenhouse gas emissions and biodiversity loss~\cite{Foley2005}. Spatial quantification of land use and land cover change allows societies to understand the extent of these impacts. Satellites are required to generate land cover products, since they provide a consistent, periodic, and~globally reaching coverage of the planet’s surface.  Thus, satellite-based land cover products are essential to support evidence-based policies that promote~sustainability.

There is currently an extensive amount of Earth observation (EO) data collected by an increasing number of satellites. Coupled with the adoption of open data policies by most spatial agencies, an~unprecedented amount of satellite data is now publicly available~\cite{Wulder2012}. This has brought a significant challenge for researchers and developers of geospatial technologies: how to design and build technologies that allow the Earth observation community to analyse big data sets?


The emergence of cloud computing services capable of storing and processing big EO data sets allows researchers to develop innovative methods for extracting information~\cite{Gorelick2017, Giuliani2019}. One of the relevant trends is to work with satellite image time series, which are calibrated and comparable measures of the same location on Earth at different times. These measures can come from a single sensor (e.g., MODIS) or by combining various sensors (e.g., Landsat 8 and Sentinel-2). When associated with frequent revisits, image time series can capture significant land use and land cover changes~\cite{Verbesselt2010}. For~this reason, developing methods to analyse image time series has become a relevant research area in remote sensing~\cite{Arvor2011, Maus2016, Pelletier2019}. 

Multiyear time series of land cover attributes enable a broader view of land change. Time series capture both gradual and abrupt changes \citep{Lambin2003}. Researchers have used time series in applications such as forest disturbance \citep{Kennedy2010}, land change \citep{Zhu2020}, ecological dynamics \citep{Pasquarella2016}, agricultural intensification \citep{Galford2008}, and~deforestation monitoring \citep{Arvor2012}. 

The traditional approach for change detection in remote sensing is to compare two classified images of the same place at different times and derive a transition matrix. \mbox{Camara~et~al.~\cite{Camara2016}} call this a space-first, time-later approach. The alternative is to adopt a time-first, space-later method, where all values of the time series are inputs for analysis. Each spatial location is associated with a time series. These algorithms first classify each time series individually and later apply spatial post-processing to capture neighbourhood information. Many authors argue that time-first, space-later methods are better suited to track changes continuously better than space-first, time-later approaches~\cite{Camara2016, Pasquarella2016, Pelletier2019,  Sudmanns2018,Woodcock2020}.

This paper describes \texttt{sits}, an~open-source R package for satellite image time series analysis using machine learning that adopts a time-first, space-later approach. Its main contribution is to provide a complete workflow for land classification of big EO data sets. Users build data cubes from images in cloud providers, retrieve time series from these cubes, and~can improve training data quality.  Different machine learning and deep learning methods are supported. Spatial smoothing methods remove outliers from the classification. Best practice accuracy techniques ensure realistic assessments. The~authors designed an expressive API that allows users to achieve good results with minimal programming effort.

The \sits package incorporates new developments in image catalogues for cloud computing services. It also includes deep learning algorithms for image time series analysis published in recent papers and not available as R packages~\cite{Pelletier2019, Fawaz2020}. The~authors developed new methods for quality control of training data~\cite{Santos2021a}. Parallel processing methods specific for data cubes ensure efficient performance.~Given these innovations, \sits provides functionalities beyond existing R packages.

We organise this paper as follows. In~Section~\ref{sec2}, we review Earth observation data cubes, pointing out the challenges involved in building them. Section~\ref{sec3} presents the design decisions for the \sits API and the internal components of the package. Section~\ref{sec4} shows a concrete example of using \sits to perform land use and land cover classification in the Brazilian Cerrado and discusses the lessons learned. We conclude by pointing out further directions in the development of the~package.

\section{Earth Observation Data~Cubes}\label{sec2}

The term Earth observation (EO) data cube is being widely used to refer to large collections of satellite images modelled as multidimensional structures to support time series analysis in an easy way to scientists~\cite{Appel2019}. There are different definitions of an EO data cube. Some authors refer to EO data cubes as organised collections of images~\cite{Lewis2017} or to the software used to produce the data collection~\cite{Giuliani2020}. Others are more restrictive, defining data cubes as regular collections reprocessed to a common projection and a consistent timeline~\cite{Ferreira2020a, Appel2019}. We propose a conceptual approach, following the idea of EO data cubes as geographical fields~\cite{Galton2004, Camara2014}. The~essential property of a geographical field is its field function; for each location within a spatiotemporal extent, this function produces a set of values. This perspective leads to the following~definitions.

\begin{Definition}\label{de1}
{A data cube is defined by a field function $f: p \to \mathbf{v}, \forall p \in ST, \exists \: \mathbf{v}$,  where $ST$ is a set of positions in space-time and $\mathbf{v}$ is a vector of attributes without missing values.}
\end{Definition}

\begin{Definition}\label{de2}
{An Earth observation data cube is a data cube whose spatiotemporal extent has a two-dimensional spatial component $S: \mathbf{X} \times \mathbf{Y}$ where $\forall p = (x_i, y_j) \in S$, the~point $p$ can be referenced to a location on the surface of the Earth, and~points in the spatial extent are mapped to a two-dimensional regular grid. 
}
\end{Definition}

\begin{Definition}\label{de3}
The temporal component of the spatiotemporal extent $ST$ is a set of time intervals $\mathbf{T} = {t_1, ..., t_n}$ such that $\forall	(i,j, i \neq j), \: t_i \cap t_j = \emptyset$ and $\forall (i, i+1), Meets(t_i, t_{i+1})$, where $Meets(.)$ is the temporal relation defined by Allen and Ferguson~\cite{Allen1994}.
\end{Definition}

\begin{figure}[H]
	\widefigure
     \includegraphics[width=0.9\textwidth]{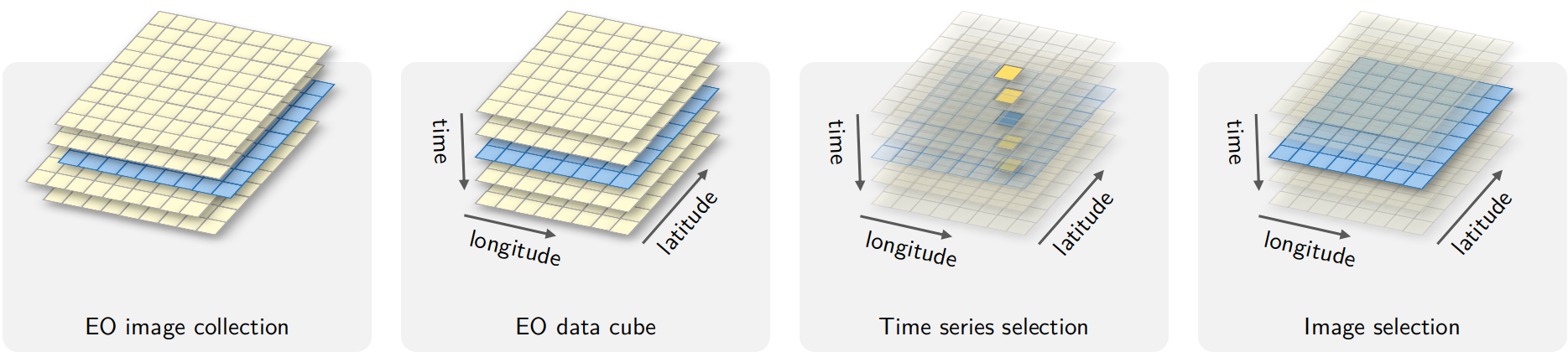}
        \caption{Conceptual view of data cubes (source: authors).}
        \label{fig:dcconcept}
\end{figure}

Definitions \ref{de1}--\ref{de3} capture the essential properties of an EO data cube: (a) there is a unique field function;  (b) the spatial support is georeferenced; (c) temporal continuity is assured; and (d) all spatiotemporal locations share the same set of attributes; {and (e) there are no gaps or missing values in the spatiotemporal extent} (See Figure~\ref{fig:dcconcept}). Since the proposed definition is an abstract one, it can be satisfied by different concrete~implementations. 
	
\end{paracol}

\begin{paracol}{2}

\switchcolumn

{EO data cubes that follow these definitions enable the use of machine learning algorithms. These methods do not allow gaps or missing values in the input data. Since image collections available in cloud services do not satisfy these requirements, such collections need additional processing. This can be done either by creating a new set of files that support the properties of a data cube, or~by developing software that creates data cubes in real-time. To~better understand the problem, consider the differences between analysis-ready data (ARD) image collections and EO data cubes.

\begin{Definition}
An ARD image collection is a set of files from a given sensor (or a combined set of sensors) that has been corrected to ensure comparable measurements between different dates. All images are reprojected to a single cartographical projection following well-established standards. Data producers usually crop ARD image collections into tiling systems.
\end{Definition}

ARD image collections do not fully support a field function, as~required by \mbox{Definition~\ref{de1}} of data cubes. These collections do not guarantee that every pixel of an image has a valid set of values since they may contain cloudy and missing pixels.  For~example, Figure~\ref{fig:imagesro} shows images of tile ``20LKP'' of the Sentinel-2/2A image collection available on the Amazon Web Service (AWS) for different dates. Some images have a significant number of clouds. To~support the data cube abstraction, data analysis software has to replace cloudy or missing pixels with valid~values.  

\begin{figure}[H]
     \includegraphics[scale=0.25]{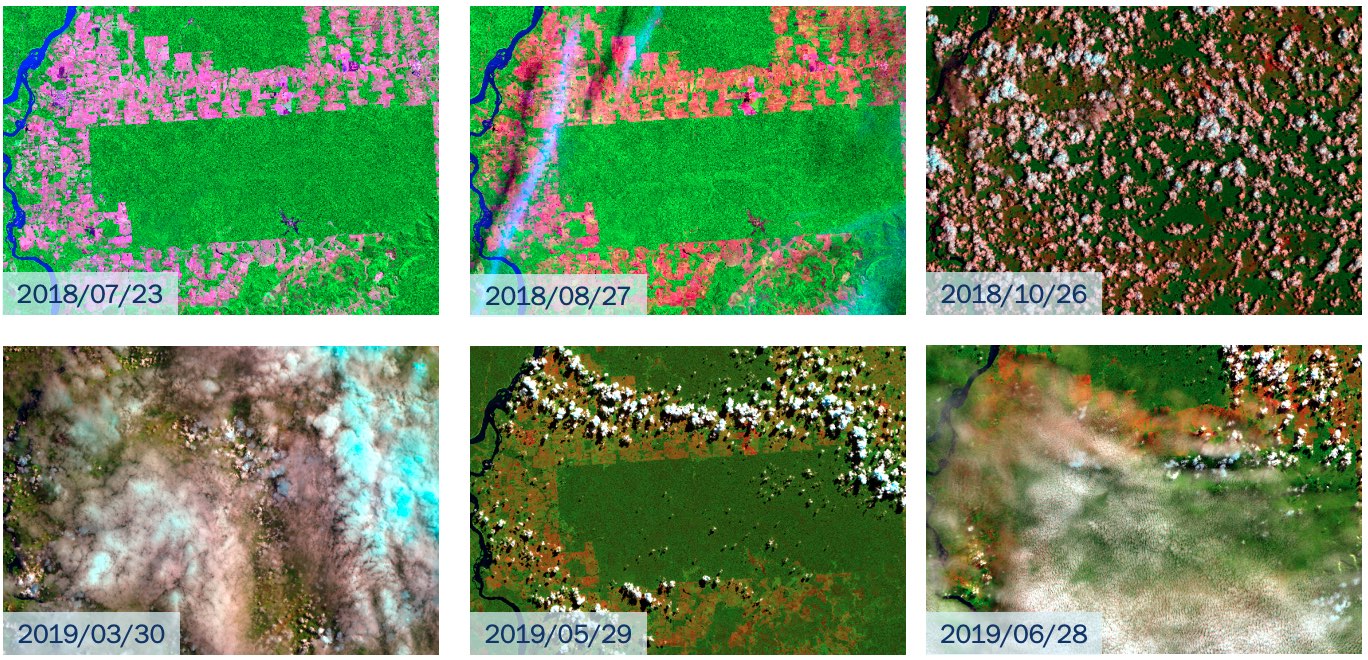}
        \caption{Sentinel-2 image colour composites for tile 20LKP on different dates (source: authors).}
        \label{fig:imagesro}
\end{figure}

A further point concerns the timeline of different tiles. Consider the neighbouring Sentinel-2 tiles ``20LLP'' and ``20LKP'' for the period 13 July 2018 to 28 July 2019. Tile 20LLP has 144 temporal instances, while tile 20LKP has only 71 instances. Such differences in temporal extent are common in large image collections. To~ensure that big areas can be processed using a single machine learning model without the need for data reprocessing, the data analysis software has to enforce a unique timeline for all tiles. 


The differences  between ARD image collections and data cubes proper have led some experts to develop tools that reprocess collections, making them regular in space and in time, and~account for missing or noisy values. For~example, the~Brazil Data Cube provides organised collections~\cite{Ferreira2020a}.~The R~\texttt{gdalcubes} package supports generating consistent data cubes~\cite{Appel2019}. When designing \texttt{sits}, the~authors decided that the software would support both kinds of imagery: regular data cubes and irregular image collections. The~design decisions for \sits will be further explored in the next~section.

\section{Software Design and Analysis~Methods}\label{sec3}
\unskip

\subsection{Requirements and Design~Choices} 

The target audience for \sits is the new generation of specialists who understand the principles of remote sensing and can write scripts in R. To~allow experts to use the full extent of available satellite imagery, \sits adopts a time-first, space-later approach. Satellite image time series are used as inputs to machine learning classifiers; the results are then post-processed using spatial smoothing. Since machine learning methods need accurate training data, \sits includes methods for quality assessment of training samples. There are no minimum requirements for spatial or temporal extents and temporal sampling frequencies for land cover classification. The software provides tools for model validation and accuracy measurements. The~package thus comprises a production environment for big EO data~analysis. 

We chose to develop \sits using the R programming environment. R is well-tested and widely used for data analysis. It has high-level abstractions for spatial data, time series, and~machine learning. R packages are community managed using a repository (CRAN) that enforces quality standards and cross-platform compatibility. Since \sits is an integrated environment that supports the full cycle of land use and land cover classification, it uses a large number of third-party packages. The~R CRAN community package management provides a sound basis for our work. Furthermore, the~authors wanted to build robust software for big EO data analysis. Since the authors of the package include experienced R developers, the~choice of R was a natural one.

Further requirements come from the authors' affiliation with Brazil's National Institute for Space Research (INPE). The~institute provides the official Brazilian estimates of deforestation and land use change in the environmentally sensitive Amazonia and Cerrado biomes. Given the emissions and biodiversity impacts of land use change in Amazonia and Cerrado, INPE has been providing estimates of deforestation since 1998. Since 2007, INPE also produces daily alerts of forest cuts~\cite{Shimabukuro2012}. Comprehensive assessments have shown the quality of INPE's work~\cite{Parente2021}. Since INPE experts aim to use \sits to generate monitoring products~\cite{Ferreira2020a}, the~package has been designed to meet the performance needs of operational~activities. 

These requirements led to the following design goals:

\begin{enumerate}
	\item {Encapsulate the land classification workflow in a concise R API.}
	\item {Provide access to data cubes and image collections available in cloud services.}
	\item {Develop methods for quality control of training data sets.}
	\item {Offer a single interface to different machine learning and deep learning algorithms.}
	\item {\textls[-10]{Support efficient processing of large areas, with~internal support for parallel processing.}}
	\item {Include innovative methods for spatial post-processing.}
\end{enumerate}

\subsection{Workflow and~API}

The design of the \texttt{sits} API considers the typical workflow for land classification using satellite image time series (see Figure~\ref{fig:sits}). Users define a data cube by selecting a subset of an ARD image collection. They obtain the training data from a set of points in the data cube whose labels are known. After~performing quality control on the training samples, users build a machine learning model and use it to classify the entire data cube. The~results go through a spatial smoothing phase that removes outliers. Thus, \sits supports the entire cycle of land use and land cover~classification. 

\begin{figure}[H]
     \includegraphics[width=0.6\textwidth]{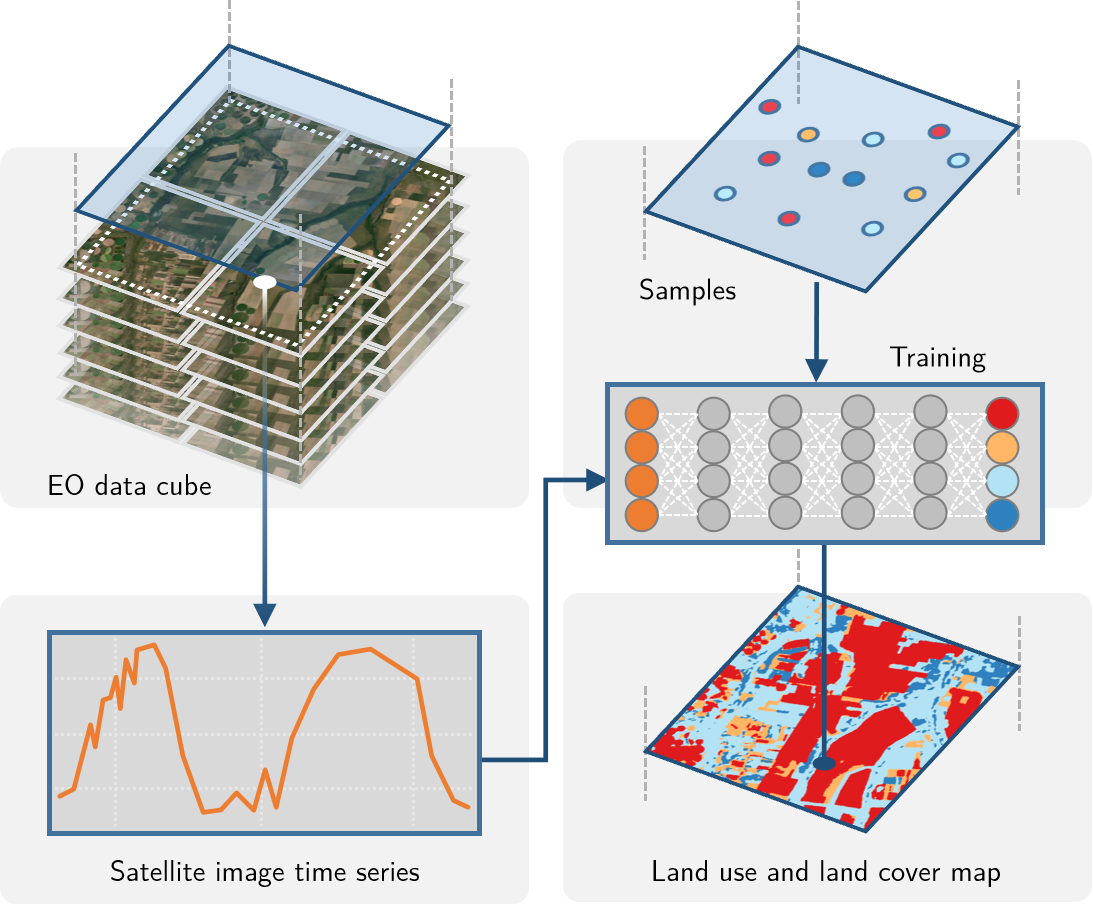}
        \caption{Using time series for land~classification (source:authors)}
        \label{fig:sits}
\end{figure}

The above-described workflow represents one complete cycle of land use classification. Machine learning methods require that the training data set and the classification input have the same number of dimensions. Thus, increasing or reducing the size of a data cube requires that the classification model be retrained. However, once a model has been trained, it can be applied to any data cube with the same dimensions. A~model trained using samples taken from a data cube can be used for classifying another data cube, provided both cubes share the same bands and the same number of temporal intervals.

When designing the \texttt{sits} API, we tried to capture the essential properties of good software. As~stated by Bloch~\cite{Bloch2006}, good APIs {``should be easy to use and hard to misuse, and~should be self-documenting''}. Bloch~\cite{Bloch2006} also recommends, {``good programs are modular, and~inter-modular boundaries define APIs''}. Following this advice, in~\texttt{sits}, each function carries out one task of the land classification workflow. For~example, instead of having separate functions for working with machine learning models, there is one function for model training. The~\texttt{sits\_train()} function encapsulates all differences between different methods, ranging from random forests to convolutional neural networks. All functions have convenient default parameters. Thus, novice users can achieve good results, while more experienced ones are able to fine-tune their models to get further~improvements. 

\textls[-35]{The \sits API captures the main steps of the workflow.~These functions are:~(a)~\texttt{sits\_cube()} creates a cube; (b) \texttt{sits\_get\_data()} extracts training data from the cube; (c) \texttt{sits\_train()} trains a machine learning model; (d) \texttt{sits\_classify()} classifies the cube and produces a probability cube; (e) \texttt{sits\_smooth()} performs spatial smoothing using the probabilities; (f) \texttt{sits\_label\_classification()} produces the final labelled image. Since these functions encapsulate the core of the package, scripts in \sits are concise and easy to reuse and reproduce.}

\subsection{Handling Data~Cubes} 

The \sits package works with ARD image collections available in different cloud services such as AWS, Microsoft, and~Digital Earth Africa. It accepts ARD image collections as input and has user-transparent internal functions that enforce the properties of data cubes (Definitions \ref{de1}--\ref{de3}). Currently, \sits supports data cubes available in the following cloud services: (a) Sentinel-2/2A level 2A images in AWS and on Microsoft's Planetary Computer; (b) collections of Sentinel, Landsat, and~CBERS images in the Brazil Data Cube (BDC); (c) collections available in Digital Earth Africa; (d) data cubes produced by the \texttt{gdalcubes} package \citep{Appel2019}; (e) local~files.

The big EO data sets available in cloud computing services are constantly being updated. For~this reason, \sits uses the STAC (SpatioTemporal Asset Catalogue) protocol. STAC is a specification of geospatial information adopted by providers of big image collections~\cite{Hanson2019}. Using STAC brings important benefits to \sits, since the software is able to access up-to-date~information through STAC end-points.

Using \texttt{sits}, the~user defines a data cube by selecting an ARD image collection and determining a space-time extent. Listing~\ref{code:cube_ex} shows the definition of a data cube using AWS Sentinel-2/2A images. The~user selects the ``Sentinel-2 Level 2'' collection in the AWS cloud service. The~data cube's geographical area is defined by the tile ``20LKP'' and the temporal extent by a start and end date. Access to other cloud services works in similar ways. Data cubes in \sits contain only metadata; access to data is done on an as-need~basis.

\begin{lstlisting}[language=R, caption = Defining a data cube in \texttt{sits}, frame = lines, label=code:cube_ex]
s2_cube <- sits_cube(
    source        = "AWS",
    name          = "T20LKP_2018_2019",
    collection    = "sentinel-s2-l2a",
    tiles         = c("20LKP","20LLP"),
    start_date    = "2018-07-18",
    end_date      = "2018-07-23",
    s2_resolution = 20
)
\end{lstlisting}

\subsection{Handling Time~Series}

Following the approach taken by the \texttt{sf} R package for handling geospatial vector objects \citep{Pebesma2018}, \sits stores time series in an object-relational table. As~shown in Figure~\ref{fig:st}, a~\sits time series table contains data and metadata. The~first six columns contain the metadata: spatial and temporal information, the~label assigned to the sample, and~the data cube from where the data have been extracted. The~spatial location is given in longitude and latitude coordinates for the WGS84 ellipsoid. For~example, the~first sample at location ($-55.2$, $-10.8$) has been labelled ``Pasture'', being valid during the interval from 14 September 2013 to 29 August 2014. The~\texttt{time series} column contains the time series data for each spatiotemporal~location.

\begin{figure}[H]
     \includegraphics[scale = 0.50]{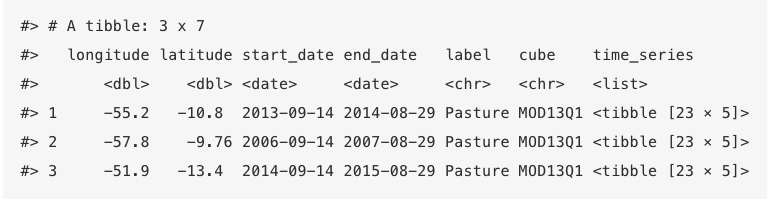}
        \caption{Data structure for time series (source: authors).}
        \label{fig:st}
\end{figure}

Time series tables store training data used for land use and land cover classification. They are built in two steps. Based on field observations or by interpreting high-resolution images, experts provide samples with valid locations, labels, and~dates. These samples can be provided as comma-separated text files or as shapefiles. Then, \sits uses the expert data to retrieve the values of time series for each location from the data cube, as~illustrated in Listing~\ref{code:csv}.

\begin{lstlisting}[language=R, caption = Extracting time series from a data cube., frame = lines, label=code:csv]
# text file containing sample information
csv_file <- "/home/user/samples.csv"
# obtain time series
samples <- sits_get_data(
    cube = s2_cube,
    file = csv_file
)
\end{lstlisting}

\subsection{Sample Quality~Control}

Experience with machine learning methods shows that the limiting factor in obtaining good results is the number and quality of training samples. Large and accurate data sets are better, no matter the algorithm used~\cite{Maxwell2018, ThanhNoi2018}, while noisy and imperfect samples have a negative effect on classification performance~\cite{Frenay2014}.  Software that uses machine learning for satellite image analysis needs good methods for sample quality~control.

The \sits package provides an innovative sample quality control technique based on self-organising maps (SOM)~\cite{Santos2019a, Santos2021}. SOM is a dimensionality reduction technique. High-dimensional data are mapped into two dimensions, keeping the topological relations between similar patterns~\cite{Kohonen1990}. The~input data for quality assessment is a set of training samples, obtained as described in the ``Handling Time Series'' subsection above. When projecting a high-dimensional data set of training samples into a 2D self-organising map, the~units of the map (called ``neurons'') compete for each sample. It is expected that good quality samples of each label should be close together in the resulting map. The~neighbours of each neuron of a SOM map provide information on intraclass and interclass~variability. 

The function \texttt{sits\_som\_map()} creates a SOM to assess the quality of the samples. Each sample is assigned to a neuron based on similarity. After~the samples are mapped to neurons, each neuron will be associated with a discrete probability distribution. Usually, homogeneous neurons (those with a single label) contain good quality samples. Heterogeneous neurons (those with two or more labels with significant probability) are likely to contain noisy samples. The~\texttt{sits\_som\_map()} function provides quality information for every sample. It also generates a 2D map that is useful to visualise class noise, since neurons associated with the same class are expected to form a cluster in the SOM~map. 

Figure~\ref{fig:sm} shows a SOM map for a set of training samples in the Brazilian Cerrado, obtained from the MODIS MOD13Q1 product. This set ranges from 2000 to 2017 and includes 50,160 land use and land cover samples divided into 12 labels: Dunes, Fallow--Cotton, Millet-Cotton, Soy--Corn, Soy--Cotton, Soy--Fallow, Pasture, Rocky Savanna, Savanna, Dense Woodland, Savanna Parkland, and~Planted Forest. Visual inspection shows several outlier neurons located far from their label cluster. For~example, while the neurons associated with the ``Pasture'' label form a cluster, some of those linked to the ``Rocky Savanna'' label are mixed among those labelled ``Dense Woodland'', an~unexpected situation. The~quantitative evaluation confirms this intuitive insight. As~shown by Santos~et~al.~\cite{Santos2021}, removing these and other outliers improves classification~results.

\begin{figure}[H]
     \includegraphics[scale = 0.50]{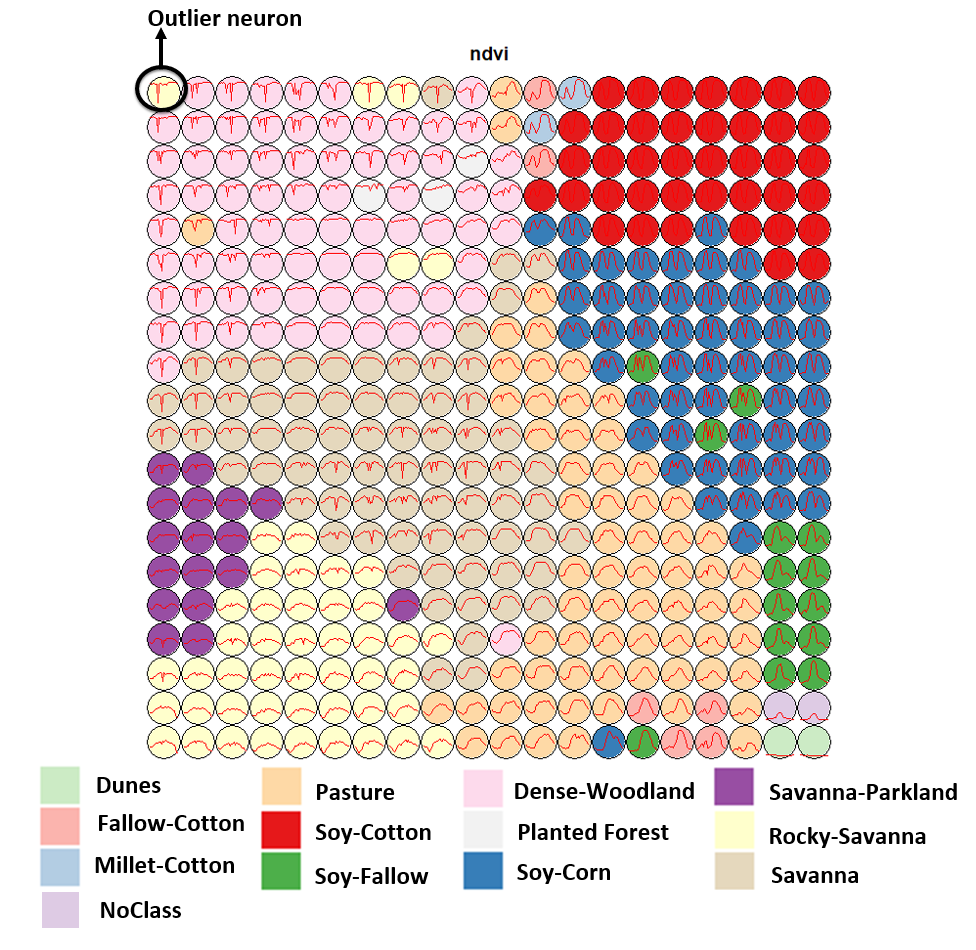}
        \caption{SOM map for Cerrado training samples (source: authors).}
        \label{fig:sm}
\end{figure}
\unskip

\subsection{Training Machine Learning~Models} 

One of the key features of machine learning and deep learning models is their dependence on the training data sets~\cite{Maxwell2018, Ma2019}. Selecting good quality training samples has a stronger impact on the accuracy of the classification maps than the choice of the machine learning method. For~this reason, the~\sits package has been designed to support users to freely choose the training data and its labels. For~each specific region of the globe and each specific aim, users select labels that match their classification schemes. Users provide samples that include geographical position, start and end dates, and~a label. The~package will then extract the associated time series from the data cube and use them as training data. There are no constraints on the choice of labels for the time series used for training~models.

After selecting good quality samples, the~next step is to train a machine learning model. The~package provides support for the classification of time series, preserving the full temporal resolution of the input data. It supports two kinds of machine learning methods. The~first group of methods does not explicitly consider spatial or temporal dimensions; these models treat time series as a vector in a high-dimensional feature space. From~this class of models, \sits includes random forests~\cite{Belgiu2016}, support  vector machines~\cite{Mountrakis2011},  extreme gradient boosting~\cite{Chen2016}, and~multi-layer perceptrons~\cite{Parente2019a}. The~authors have used these methods with success for classifying large areas~\cite{Picoli2018, Picoli2020a, Ferreira2020a}. Our results show that, given good quality samples, \sits can achieve high classification accuracy using feature space machine learning~models.

The second group of models comprises deep learning methods designed to work with image time series. Temporal relations between observed values in a time series are taken into account.  Time series classification models for satellite data include 1D convolution neural networks (1D-CNN)~\cite{Pelletier2019, Fawaz2020}, recurrent neural networks (RNN)~\cite{Russwurm2018}, and~attention-based deep learning~\cite{Garnot2020a, Russwurm2020}. The~\sits package supports a set of 1D-CNN algorithms: TempCNN~\cite{Pelletier2019}, ResNet~\cite{Fawaz2019}, and~InceptionTime~\cite{Fawaz2020}. Models based on 1D-CNN treat each band of an image time separately. The~order of the samples in the time series is relevant for the classifier. Each layer of the network applies a convolution filter to the output of the previous layer. This cascade of convolutions captures time series features in different time scales~\cite{Pelletier2019}. In~the Results section of the paper, we show the use of a TempCNN model to classify the Cerrado biome in Brazil for the year~2018. 

{Since \sits is aimed at remote sensing users who are not machine learning experts, the~package provides a set of default values for all classification models. These settings have been chosen based on extensive testing by the authors. Nevertheless, users can control all parameters for each model. The~package documentation describes in detail the tuning parameters for all models that are available in the respective functions. Thus, novice users can rely on the default values, while experienced ones can fine-tune model parameters to meet their needs.}

\subsection{Data Cube~Classification} 

The \sits package runs in any computing environment that supports R. When working with big EO data, the~target environment for \sits is a virtual machine located close to the data repository. To~achieve efficiency, \sits implements its own parallel processing. Users are not burdened with the need to learn how to do multiprocessing and, thus, their learning curve is~shortened.

Memory management in R is a hard challenge. Some advanced machine learning and deep learning methods require dedicated environments outside \R. For~example, deep learning methods in \sits use the~\texttt{keras} R package. In turn, this package calls Python code that provides a front-end to the C++ TensorFlow library. 
All of these dependencies cause R  to not have a predictable memory allocation behaviour when doing parallel processing. Given this situation, we developed a customised parallel processing implementation for \sits to work well with big EO~data.

After many tests with different R packages that provide support for parallel processing, we found out that no current R package meets our needs. The~authors implemented a new fault tolerant multi-tasking procedure for big EO data classification. Image classification in \sits is done by a cluster of independent workers linked to one or more virtual machines. To~avoid communication overhead, all large payloads are read and stored independently; direct interaction between the main process and the workers is kept at a minimum. The~customised approach is depicted in Figure~\ref{fig:sits_parallel}.

\begin{enumerate}
    \item {Based on the size of the cube, the~number of cores, and~the available memory, divide the cube into chunks.}
    \item {{The cube is divided into chunks along its spatial dimensions. Each chunk contains all temporal intervals.}}
    \item {Assign chunks to the worker cores. Each core processes a block and produces an output image that is a subset of the result.}
    \item {After all the subimages are produced, join them to obtain the result.}
    \item {If a worker fails to process a block, provide failure recovery and ensure the worker completes the job.}
\end{enumerate}

This approach has many advantages. It works in any virtual machine that supports R and has no dependencies on proprietary software. Processing is done in a concurrent and independent way, with~no communication between workers. Failure of one worker does not cause failure of the big data processing. The~software is prepared to resume classification processing from the last processed chunk, preventing against failures such as memory exhaustion, power supply interruption, or~network breakdown. From~an end-user point of view, all work is done smoothly and~transparently.

\begin{figure}[H]
     \includegraphics[width=0.6\textwidth]{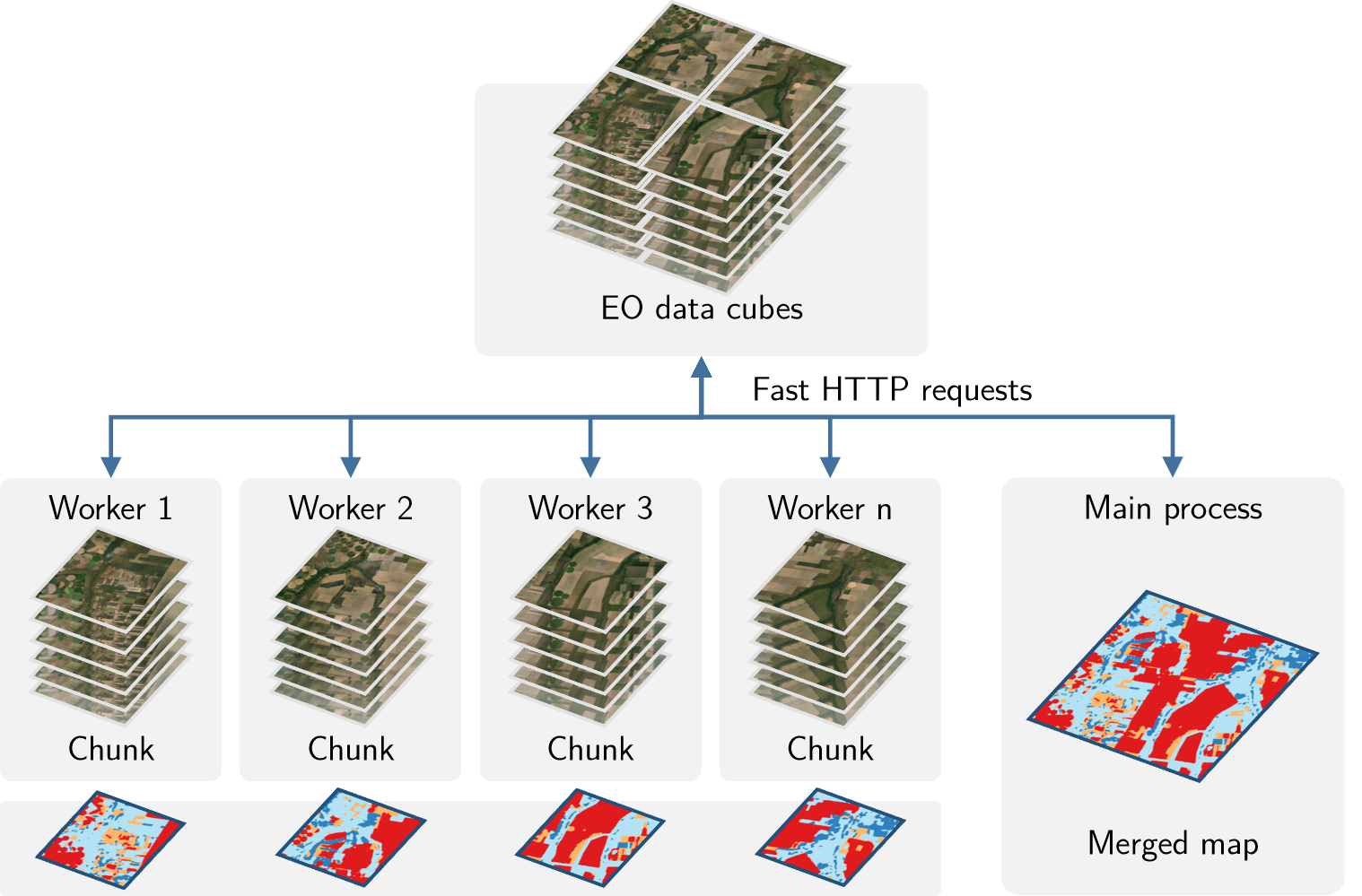}
        \caption{Parallel processing in \sits (source: authors).}
        \label{fig:sits_parallel}
\end{figure}
\unskip

\subsection{Post-Processing}

When working with big EO data sets, there is a considerable degree of data variability in each class. As~a result, some pixels will be misclassified. These errors are more likely to occur in transition areas between classes or when dealing with mixed pixels. To~offset these problems, \sits includes a post-processing smoothing method based on Bayesian probability that uses information from a pixel's neighbourhood to reduce uncertainty about its~label.

The post-classification smoothing uses the output probabilities of a machine learning algorithm. Generally, we label a pixel $p_i$ as being of class $k$ if the probability of that pixel belonging to class $k$ is higher than any other probability associated with the pixel. Instead of using these probabilities directly, Bayesian smoothing first performs a mathematical transformation by taking the log of the odds ratio for each pixel.
\begin{equation}
    \mathbf{x_i} = \log [p_{i,k} / (1-p_{i,k})]
\end{equation}

To allow mathematical tractability, we assume that $\mathbf{x_i}$ follows a multivariate normal distribution $\mathcal{N}_{k}(\boldsymbol\theta_{i},\boldsymbol\Sigma_{i})$ where $k$ is the number of classes. This distribution has  an unknown mean $\boldsymbol\theta_i$ and an estimated {a priori} covariance matrix $\boldsymbol\Sigma_{i}$ that controls the level of smoothness to be applied. The~covariance matrix represents our prior belief in the class variability and possible confusion between~classes. 

The local uncertainty is modelled by a multivariate normal distribution $\mathcal{N}_{k}(\mathbf{m}_i,\mathbf{S}_i)$ where $k$ is the number of classes. The~distribution has mean $\mathbf{m}_i$ and covariance matrix $\mathbf{S}_i$. Our strategy to reduce local uncertainty is to estimate these parameters from the neighbourhood of pixel $p_i$. Taking $\mathbf{X_i}$ as the set of all $\mathbf{x_i}$ vectors in a neighbourhood, we compute $\mathbf{m_i}=\operatorname{E}[\mathbf{X_i}]$ and $\mathbf{S}_i=\operatorname{cov}[\mathbf{X_i},\mathbf{X_i}]$. The~point estimator $\hat{\boldsymbol\theta_i}$ for each pixel $p_i$ that minimises the quadratic loss functions can be obtained by applying Bayes' rule. The~posterior estimator for the pixel's probabilities can be expressed as
\begin{equation}
    \hat{\boldsymbol{\theta_i}}=\operatorname{E}[\boldsymbol\theta_i|\mathbf{x_i}]=\boldsymbol\Sigma_i\left(\boldsymbol\Sigma_i+\mathbf{S_i}\right)^{-1}\mathbf{m_i} +
    \mathbf{S_i}\left(\boldsymbol\Sigma_i+\mathbf{S_i}\right)^{-1}\mathbf{x_i}.
\end{equation}

This estimator is computed for each pixel, producing a smoothed map. It is a weighted combination of $\mathbf{x}_i$ and the neighbourhood mean $\mathbf{m}_i$, where the weights are determined by the covariance matrices $\boldsymbol\Sigma_i$ and $\mathbf{S_i}$. The~component $\left(\boldsymbol\Sigma_i+\mathbf{S_i}\right)^{-1}$ plays a normalisation role. Given that the smoothing factor $\boldsymbol\Sigma_i$ is provided {a priori} by the user, the~estimate depends only on the neighbourhood covariance matrix $\mathbf{S_i}$. When the $\mathbf{X}$ values in a neighbourhood of a pixel are similar, the~matrix $\mathbf{S_i}$ increases relative to $\boldsymbol\Sigma_i$. In~this case, we will have more confidence in the original pixel value and less confidence in the neighbourhood mean $\mathbf{m}$. Likewise, when the $\mathbf{X}$ values in a neighbourhood of a pixel are diverse, the~values of the correlation matrix will be low. Thus, the~weight expressed by $\mathbf{S_i}$ will decrease relative to $\boldsymbol\Sigma_i$. We will have less confidence in the original pixel value $\mathbf{x_i}$ and more confidence in the local mean $\mathbf{m_i}$. The~smoothing procedure is thus most relevant in~situations where the original classification odds ratio is low, indicating a low level of separability between classes. In~these cases, the~updated values of the classes will be influenced by the local class variance. The~resulting smoothed map will thus consider the influence of the neighbours only when the confidence in the most likely label for a pixel is low. {Bayesian smoothing is an established technique for handling outliers in spatial data \citep{Cressie1995}. Its application in \sits is useful to incorporate spatial effects in the result of time series classification.}

\subsection{Validation and Accuracy~Assessment}

The \sits package offers support for cross-validation of training models and accuracy assessment of results. Cross-validation estimates the expected prediction error. It uses part of the available samples to fit the classification model, and~a different part to test it. The~\sits software performs  \textit{k-fold} validation. The~data are split into $k$ partitions with approximately the same size. The~model is tested $k$ times. At~each step, \sits takes one distinct partition for testing and the remaining ${k-1}$ partitions for training the model. The~results are averaged to estimate the prediction error. The~estimates provided by validation are a ``best-case'' scenario, since they only use  the training samples, which are subject to selection bias. Thus, validation is best used to compare different models for the same training data. Such results must not be used as accuracy~measures. 

To measure the accuracy of classified images, \sits provides a  function that calculates area-weighted estimates~\cite{Olofsson2013, Olofsson2014}. The~need for area-weighted estimates arises because land use and land cover classes are not evenly distributed in space. In~some applications (e.g., deforestation) where the interest lies in assessing how much has changed, the~area mapped as deforested is likely to be a small fraction of the total area. If~users disregard the relative importance of small areas where change is taking place, the~overall accuracy estimate will be inflated and unrealistic. For~this reason, the~\texttt{sits\_accuracy\_area()} function adjusts the mapped areas to eliminate bias resulting from classification error. This function provides error-adjusted area estimates with confidence intervals, following the best practices proposed by Olofsson~et~al.~\cite{Olofsson2013, Olofsson2014}.

\subsection{Extensibility}

Since one of the design aims of \sits is to keep a simple application programming interface, it uses the R S3 object model, which is easily extensible. The~designers gave particular attention to the support required for machine learning researchers to include new models in \texttt{sits}. For~machine learning models, \sits uses two R constructs.~In R, classifiers should provide a \texttt{predict} function, which carries out the actual assignment of input to class probabilities. R also provides support for closures, which are functions written by functions~\cite{Wickham2019}. Using closures is particularly useful for dealing with machine learning functions that have completely different internal implementations. The~\texttt{sits\_train()} function in \sits is a closure that encapsulates the details of how the classifier works. The~closure returns a function to classify time series and data cubes using the overloaded R \texttt{predict} function. Therefore, training and classification in \sits are independent and extensible. Users can provide new models without any need for changing the other components of the package. We expect that both the authors and other contributors to the package will include further advanced models tailored for image time~series.

Furthermore, \sits can be used together with other R packages or integrated with different programming languages. It can be used to prepare time series for other algorithms, since time series in \sits use a tabular format easily exportable to open access formats. Images associated with a data cube can also be exported from cloud providers to local repositories. Users can save deep learning classification models in TensorFlow format for later processing. Python programmers can access the full \sits API through interface packages such as \texttt{rpy2}. This broadens the potential community of contributors and users of \texttt{sits}.

\section{Results and~Discussion}\label{sec4}

{In this section, we present an application of \sits to produce a one-year land use and cover classification of the Cerrado biome in Brazil using Landsat-8 images. Cerrado is the second largest biome in Brazil with 1.9 million km$^2$.  The~Brazilian Cerrado is a tropical savanna ecoregion with a rich ecosystem ranging from grasslands to woodlands. It is home to more than 7000 species of plants with high levels of endemism \citep{Klink2005}. It includes three major types of natural vegetation: \emph{Open Cerrado}, typically composed of grasses and small shrubs with a sporadic presence of small tree vegetation; \emph{Cerrado Sensu Stricto}, a~typical savanna formation, characterised by the presence of low, irregularly branched, thin-trunked trees; and \emph{Cerradão}, a~dry forest of medium-sized trees (up to 10--12 m)~\cite{Goodland1971, Del-Claro2019}. Its natural areas are being converted to agriculture at a fast pace, as~it is one of the world's fast moving agricultural frontiers \citep{Walter2006}. The~main agricultural land uses include cattle ranching, crop farms, and~planted~forests.}

\subsection{Input~Data}

The Brazilian Cerrado is covered by 51 Landsat-8 tiles available in the Brazil Data Cube (BDC)~\cite{Ferreira2020}. Each Landsat tile in the BDC covers a 3\degree $\times$ 2\degree grid in Albers equal area projection with an area of 73,920 km$^2$, and a size of \mbox{11,204 $\times$ 7324 pixels}. The~one-year classification period ranges from September 2017 to August 2018, following the agricultural calendar. The~temporal interval is 16 days, resulting in 24 images per tile. We use seven spectral bands plus two vegetation indexes (NDVI and EVI) and the cloud cover information. The~total input data size is about 8~TB.  

\subsection{Training~Samples}

{Since the Cerrado is Brazil's main agricultural frontier, our classification aims to identify both natural vegetation and agricultural lands. Its large latitude gradient includes different climate regimes, which lead to important differences in the spectral responses of land cover types. To~classify the biome with good accuracy, one needs a large and good quality sample set. To~obtain good training data, we carried out a systematic sampling using a grid of 5 $\times$ 5 km throughout the Cerrado biome, collecting 85,026 samples. The~training data labels were extracted from three sources: the pastureland map of 2018 from Pastagem.org \citep{Parente2019}, MapBiomas Collection 5 for  2018 \citep{Souza2020}, and~Brazil's National Mapping Agency IBGE maps for 2016--2018 \citep{IBGE2020}. Out of the 85,026 samples, we selected those where there was no disagreement between the labels assigned by the three sources. The~resulting set had 48,850 points from which we extracted the time series using the Landsat-8 data cube. The~distribution of samples for each class is the following: ``Annual Crop'' (6887), ``Cerradao'' (4211), ``Cerrado'' (16,251), ``Natural Non Vegetated'' (38), ``Open Cerrado'' (5658), ``Pasture'' (12,894), ``Perennial Crop'' (68), ``Silviculture'' (805), ``Sugarcane'' (1775), and~``Water'' (263). The~effort to obtain representative samples is justified by the importance of training data in the classification results.} 

\subsection{Training and~Classification}

{The data set of 48,850 samples was used to train a convolutional neural network model using the TempCNN method \citep{Pelletier2019}. All available attributes in the BDC Landsat-8 data cube (two vegetation indices and seven spectral bands) were used for training and classification. The~\sits commands are illustrated in Listing \ref{code:cerrado_model}. We used the default configuration of the TempCNN method with three 1D convolutional layers~\cite{Pelletier2019}. After~the classification, we applied Bayesian smoothing to the probability maps and then generated a labelled map by selecting the most likely class for each pixel. The~classification was executed on an Ubuntu server with 24 cores and 128 GB memory. Each Landsat-8 tile was classified in an average of 30 min, and~the total classification took about 24 h. Figure~\ref{fig:cerrado_lulc} shows the final map.}

\begin{figure}[H]
     \includegraphics[width=0.7\textwidth]{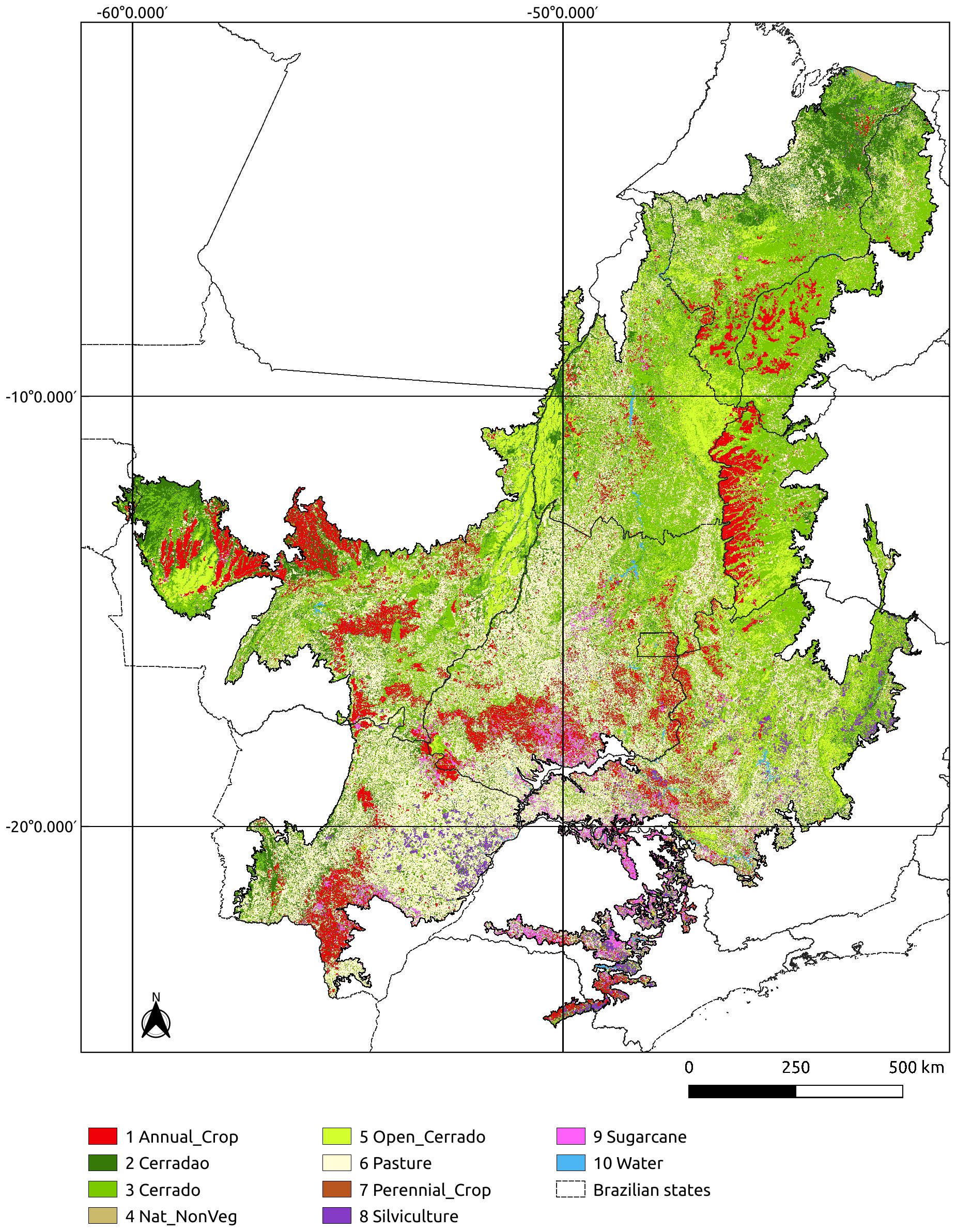}
        \caption{Cerrado land use and land cover map for 2018 (source: authors.}
    \label{fig:cerrado_lulc}
\end{figure}
\unskip

\subsection{Code}

According to the principles of \sits API, each function is responsible for one of the workflow tasks. As~a consequence, the~classification of the whole Cerrado (200 million ha) is achieved by six R commands, as~shown in Listing \ref{code:cerrado_model}. 

\begin{lstlisting}[language=R, caption= R Code for Cerrado classification., label=code:cerrado_model, frame=lines]
# define a reference to a data cube stored in the BDC  using
# a shapefile with Cerrado boundaries as a region of interest
l8_cerrado_cube <- sits_cube(
    source     = "BDC",
    name       = "cerrado",
    collection = "LC8_30_16D_STK-1",
    roi        = "./cerrado.shp",
    start_date = "2017-09-01",
    end_date   = "2018-08-31"
)
# obtain the time series for the samples
# input data is a  CSV file
samples_cerrado_lc8 <- sits_get_data(
    data = cube, 
    file = "./samples_48K.csv"
)
# train model using TempCNN algorithm
cnn_model <- sits_train(
    data      = samples_cerrado_lc8, 
    ml_method = sits_TempCNN()
)
# classify data cube
probs_cerrado <- sits_classify(
    data     = l8_cerrado_cube,
    ml_model = cnn_model
)
# compute Bayesian smoothing
probs_smooth <- sits_smooth(
    data = probs_cerrado
)
# generate thematic map
map <- sits_label_classification(
    data = probs_smooth
)
\end{lstlisting}

\subsection{Classification~Accuracy}

{To obtain an accuracy assessment of the classification, we did a systematic sampling of the Cerrado biome using a 20 $\times$ 20 km grid with a total of 5402 points.} These samples are independent of the training set used in the classification. They were interpreted by five specialists using high resolution images from the same period of the classification. For~the assessment, we merged the labels ``Cerradao'', ``Cerrado'', and~``Open Cerrado'' into one label called ``Cerrado''. We also did additional sampling to reach a minimal number of samples for the classes ``Natural Non Vegetated'', ``Perennial Crop'', and~``Water''. This resulted in 5286 evaluation samples thus distributed: ``Annual Crop'' (553), ``Cerradao'' (704), ``Cerrado'' (2451), ``Natural Non Vegetated'' (44), ``Pasture'' (1246), ``Perennial Crop'' (38), ``Silviculture'' (94), ``Sugarcane'' (77), and~``Water'' (79). We used the \sits implementation of the area-weighted technique~\cite{Olofsson2013} to provide an unbiased estimator for the overall accuracy and the total area of each class based on the reference samples. The~classification accuracies are shown in Table~\ref{tab:kfold_validation}. The~overall accuracy of the classification was~0.86.

\begin{specialtable}[H] 
\caption{Area-weighted classification~accuracy.}
\label{tab:kfold_validation}
\setlength{\cellWidtha}{\columnwidth/3-2\tabcolsep-0.0in}
\setlength{\cellWidthb}{\columnwidth/3-2\tabcolsep+0.0in}
\setlength{\cellWidthc}{\columnwidth/3-2\tabcolsep+0.0in}
\scalebox{1}[1]{\begin{tabularx}{\columnwidth}{>{\PreserveBackslash\raggedright}m{\cellWidtha}>{\PreserveBackslash\centering}m{\cellWidthb}>{\PreserveBackslash\centering}m{\cellWidthc}}
\toprule
\textbf{Labels}          & \textbf{Producer's Accuracy} & \textbf{User's Accuracy} \\ \midrule
Annual Crop              & 0.81      & 0.88     \\
Cerrado                  & 0.89      & 0.91     \\
Natural Non Vegetated    & 0.63      & 0.95     \\ 
Pasture                  & 0.82      & 0.76     \\
Perennial Crop           & 0.51      & 0.74     \\
Silviculture             & 0.83      & 0.91     \\
Sugarcane                & 0.96      & 0.81     \\
Water                    & 0.93      & 0.97    \\ 
\bottomrule

\end{tabularx}}

{\footnotesize Overall Accuracy: 0.86.}

\end{specialtable}

\subsection{Discussion of~Results}

The good results of mapping the entire Cerrado biome using \sits show that the software has met its design goals. It was possible to classify a large area of about 200~million ha using an advanced deep learning model on time series with good performance. Running the whole training and classification process requires a script with only six R commands. No specific knowledge of parallel processing was required. All in all, the~concept of having an integrated solution has been~demonstrated.

The classes that have the worst performance are ``Perennial Crop'' and ``Natural Non Vegetated'' with producer's accuracy of 51\% and 63\%, respectively. Since these classes are associated with small areas of the Cerrado biome, they had fewer training samples. The~authors had access to only 68 samples of the ``Perennial Crop'' class and 38~samples of the ``Natural Non Vegetated'' class. To~improve the accuracy of these classes, future classifications need to ensure that the number of samples per class is balanced and~representative.

\section{Comparative~Analysis}\label{sec5}

{Considering the aims and design of \sits, it is relevant to discuss how its design and implementation choices differ from other open-source algorithms for the analysis of image time series. Such algorithms include BFAST for detecting trends and breaks~\cite{Verbesselt2010}, CCDC for continuous change detection~\cite{Zhu2014}, and~Time-Weighted Dynamic Time Warping (TWDTW) for land use and land cover classification~\cite{Maus2019}. These methods take a multiyear time series and break it into segments, which are then classified. In~general, time series segments have different temporal extents. Classified pixels will have one label for each segment of the associated time series.  BFAST detects trends and seasonal components in time series and has been used successfully to detect deforestation and forest degradation~\cite{Hamunyela2016}. CCDC decomposes the time series using spherical harmonics, extracting metrics for each segment; these metrics are used by random forest models to obtain land cover maps~\cite{Arevalo2020}. CCDC and BFAST are adaptive  algorithms, which can be updated as new observations become available. TWDTW uses a modified version of the dynamic time warping (DTW) distance to match time series segments to predefined patterns. The~TWDTW algorithm has been used for deforestation and cropland mapping~\cite{Cheng2019, Belgiu2018}. Thus, BFAST, CCDC, and~TWDTW are segmentation-based algorithms that differ in the ways to find breaks in the time series.}

In contrast to segmentation-based methods, \sits uses time series of predefined sizes for classification, typically covering one year. Instead of extracting metrics from time series segments, \sits uses all values of the time series. There are two types of classifiers: (a) feature space methods, where all inputs are treated equally as part of an n-dimensional space, including SVM, Random Forest, and~multilayer perceptrons; and (b) multiresolution methods based on 1D convolutional neural networks, including TempCNN and \mbox{ResNet~\cite{Pelletier2019, Fawaz2020}}. The~authors are not aware of benchmarks comparing segmentation-based methods with those based on feature space or multiresolution. As~the field of satellite image time series analysis matures, users will find enough evidence to choose which method best fits their~needs.

In what follows, we also compare \sits to other approaches for big EO data analytics, such as Google Earth Engine~\cite{Gorelick2017}, Open Data Cube~\cite{Lewis2017} and openEO~\cite{Schramm2021}. 

Google Earth Engine (GEE)~\cite{Gorelick2017} uses the Google distributed file system~\cite{Ghemawat2003} and its map-reduce paradigm~\cite{Dean2008}. By~combining a flexible API with an efficient back-end processing, GEE has become a widely used platform~\cite{Amani2020}. To use deep learning models in GEE, users have to develop, train and run these models outside of the Earth Engine in Google's AI platform. Since Google's AI platform does not have ready-to-use models for satellite image time series, users have to develop models for EO data analysis using the TensorFlow API. Doing so requires specialised knowledge outside of the scope of most remote sensing experts. Images also need to be exported from GEE to the AI platform, a~task that can be cumbersome for large data sets. By~contrast, \sits provides support for deep learning models that have been tested and validated in the scientific literature \citep{Pelletier2019, Fawaz2020}. These models are available directly in the \sits API; users do not need to understand TensorFlow to apply them.  Thus, currently it is easier to use \sits than GEE for running deep learning models in image time series.

The Open Data Cube (ODC) is an important contribution to the EO community and has proven its usefulness in many domains~\cite{Lewis2017, Giuliani2020}. It reads subsets of image collections and makes them available to users as a Python \texttt{xarray} structure. ODC does not provide an API to work with \texttt{xarrays}, relying on the tools available in Python. This choice allows much flexibility at the cost of increasing the learning curve. It also means that temporal continuity is restricted to the \texttt{xarray} memory data structure; cases where tiles from an image collection have different timelines are not handled by ODC. The~design of \sits takes a different approach, favouring a simple API with few commands to reduce the learning curve. Processing and handling large image collections in \sits does not require knowledge of parallel programming tools. Thus, \sits and ODC have different aims and will appeal to different classes of~users. 

Designers of the openEO API~\cite{Schramm2021} aim to support applications that are both language-independent and server-independent. To~achieve their goals, openEO designers use microservices based on REST protocols. The~main abstraction of openEO is a process, defined as an operation that performs a specific task. Processes are described in JSON and can be chained in process graphs. The~software relies on server-specific implementations that translate an openEO process graph into an executable script. Arguably, openEO is the most ambitious solution for reproducibility across different EO data cubes. To~achieve its goals, openEO needs to overcome some challenges. Most data analysis functions are not self-contained. For~example, machine learning algorithms depend on libraries such as TensorFlow and Torch. If~these libraries are not available in the target environment, the~user-requested process may not be executable. Thus, while the authors expect openEO to evolve into a widely used API, it is not yet feasible to base a user-driven operational software such as \sits in openEO. 

Producing software for big Earth observation data analysis requires making compromises between flexibility, interoperability, efficiency, and~ease of use. GEE is constrained by the Google environment and excels at certain tasks (e.g., pixel-based processing) while being limited in others such as deep learning. ODC allows users to complete flexibility in the Python ecosystem, at~the cost of limitations when working with large areas and requiring programming skills. The~openEO API achieves platform independence but needs additional effort in designing drivers for specific languages and cloud services. While the \sits API provides a simple and powerful environment for land classification, it has currently no support for other kinds of EO applications. Therefore, each of these solutions has benefits and drawbacks. Potential users need to understand the design choices and constraints to decide which software best meets their~needs.

\section{Conclusions} \label{sec6}

The development of analytical software for big EO data faces several challenges. Designers need to balance between conflicting factors. Solutions that are efficient for specific hardware architectures can not be used in other environments. Packages that work on generic hardware and open standards will not have the same performance as dedicated solutions. Software that assumes that its users are computer programmers are flexible, but~may be difficult for a wide audience to learn. The~challenges lead a diversity of solutions in academia and industry to work with big Earth observation data. Arguably, it is unlikely that a single approach will emerge as the complete best solution for big EO~analytics.

Despite the challenges, there are points of convergence and common ground between most of the solutions for big EO data. The~STAC protocol has emerged as a {de facto} standard for describing EO image collections. Users need interoperable and reusable solutions, where the same software can be used in different cloud services with similar results. Experience with existing solutions shows the benefits of simple APIs for the remote sensing community at large. These commonalities should be considered by big EO software~designers. 

In the design of \texttt{sits}, we had to make choices. We made an early choice to focus on time series analysis, based on the hypothesis that time series provide an adequate description of changes in land use and land cover. Instead of relying on time series metrics, we opted to allow machine learning methods to find patterns in multidimensional spaces by providing them all with available data. The~design of the \sits data structures and API follows from our choice of performing land classification based on time series~analysis. 

The limitations of \sits should also be considered. As~discussed previously, classification in \sits uses fixed time intervals. This is a convenient choice for working with machine learning, but~reduces its power to monitor continuous change. In~cases where users want to monitor subtle changes such as forest degradation, segmentation-based methods can in principle provide more detailed information. Moreover, \sits is pixel-based; each time series is associated with a pixel. Recent works show that object-based time series analysis can perform better than pixel-based approaches~\cite{Belgiu2018}. A~further issue is the need to convert ARD image collections into data cubes to work with \texttt{sits}, as discussed in Section~\ref{sec2}. Despite these limitations, \sits provides a simple API for all steps of land classification using satellite image time series. 

Plans for evolution of the \sits package include improvements to the classification workflow. We plan to include manipulation of data cubes, allowing mathematical operations to be performed. Another priority is improving the training phase, using techniques such as active learning and semi-supervised learning. Moreover, we are investigating an extension of \sits to work with spatial objects as well as including support for new classifiers. We also intend to include rule-based post-processing to allow multiyear classification comparison. Given the global applicability of \texttt{sits}, we intend to support the user and developer community by providing guidance and~documentation.


\vspace{6pt} 

\authorcontributions{Conceptualization, G.C.; data curation, R.S.; formal analysis, R.S. and G.C.; funding acquisition, G.Q. and K.F.; investigation, R.S. and G.C.; methodology, R.S., G.C., L.S., A.C. and K.F.; resources, G.Q.; software development, R.S., G.C., G.Q., L.S., F.S. and P.R.A.; validation, R.S.; visualization, R.S., G.C., F.S. and L.S.; writing---original draft, R.S. and G.C.; writing---review and editing, G.Q., P.R.A., L.S. and K.F. All authors have read and agreed to the published version of the manuscript.}

\funding{This research was supported by the Amazon Fund through the financial collaboration of the Brazilian Development Bank (BNDES) and the Foundation for Science, Technology and Space Applications (FUNCATE), process 17.2.0536.1. We also acknowledge support from  Coordenação de Aperfeiçoamento de Pessoal de Nível Superior-Brasil (CAPES) and from the Conselho Nacional de Desenvolvimento Científico e Tecnológico (CNPq) under grant 140684/2016-6 (R.S.). Additional funding was provided by the São Paulo State Foundation (FAPESP) under eScience Program grant 2014/08398-6. This work was also supported by the International Climate Initiative of the Germany Federal Ministry for the Environment, Nature Conservation, Building and Nuclear Safety (IKI) under grant 17-III-084- Global-A-RESTORE+ (“RESTORE+: Addressing Landscape Restoration on Degraded Land in Indonesia and Brazil”).}

\dataavailability{The~data and code scripts that support the findings of this study are openly available on GitHub at \url{https://github.com/e-sensing/sitsdata}, accessed on 14 June 2021.} 

\acknowledgments{The authors thank Marius Appel for the R package \texttt{gdalcubes} used in \texttt{sits}, Charlotte Pelletier and~Hassan Fawaz for openly sharing the python code that has been reused for the TempCNN and ResNet deep learning  models, and~Michelle Picoli, Anderson Soares, Michel Chaves, and~Alber Sánchez for collecting the validation data~set.}

\conflictsofinterest{The authors declare no conflict of~interest.}

\codeavailability{The \sits package is available on Github at \url{https://github.com/e-sensing/sits}, accessed on 14 June 2021. The~software is licensed under the GNU General Public License v3.0. Full documentation of the package is available at \url{https://e-sensing.github.io/sitsbook/}, accessed on 14 June 2021.}

}
\end{paracol}
\reftitle{References}

\externalbibliography{yes}

\end{document}